\documentclass{article}

    \usepackage[preprint]{neurips_2020}

\usepackage[utf8]{inputenc} 
\usepackage[T1]{fontenc}    
\usepackage{hyperref}       
\usepackage{url}            
\usepackage{booktabs}       
\usepackage{amsfonts}       
\usepackage{nicefrac}       
\usepackage{microtype}      
\usepackage{graphicx}
\usepackage{algorithm}
\usepackage{algorithmic}

\title{Effectiveness of Optimization Algorithms in Deep Image Classification}

\author{%
    {Zhaoyang Zhu$^{1*}$ \ \ Haozhe Sun$^{2*}$ \ \ Chi Zhang$^{3*}$}\\
    $^1$Department of Mechanical and Industrial Engineering\\ $^2$Department of Electrical and Computer Engineering\\ $^3$Department of Computer Science\\
    \texttt{\{zhaoyang.zhu, sunhaozhe.sun, matthewchi.zhang\}@mail.utoronto.ca} \\
}

\begin{document}

\maketitle
\begin{abstract}

Adam[4] is applied widely to train neural networks. Different kinds of Adam methods with different features pop out. Recently two new adam optimizers, AdaBelief[1] and Padam[5] are introduced among the community. We analyze these two adam optimizers and compare them with other conventional optimizers (Adam, SGD + Momentum) in the scenario of image classification. We evaluate the performance of these optimization algorithms on AlexNet[8] and simplified versions of VGGNet[7], ResNet[9] using the EMNIST[6] dataset. (Benchmark algorithm is available at \hyperref[https://github.com/chuiyunjun/projectCSC413]{https://github.com/chuiyunjun/projectCSC413}). 

\end{abstract}
\section{Introduction}
Modern deep neural networks often have a huge amount of trainable parameters, especially in complex image classification and recognition tasks. For instance, AlexNet[8] has 62 million trainable parameters, ResNet-50[9] has over 23 million trainable parameters, VGGNet[7] has 138 million trainable parameters. Thus, a fast, effective optimization algorithm for learning those parameters will not only improve model performance but also greatly reduce the cost of time and money.

Adam[4] is a stochastic optimization algorithm applied widely to train deep neural networks,  it has the advantages of RMSProp[10], Momentum, and incorporates adaptive learning rate for learning different parameters. Recently, AdaBelief[1] and Padam[5] are introduced among the community. These two algorithms are proposed to improve the performance of the original Adam optimizer in certain situations.

In this study, we analyze these two newly proposed algorithms and compare them with traditional methods Adam, SGD with momentum. We evaluate these 4 different optimizer algorithms by respectively applying each of them into a simplified version of AlexNet,VGGNet, ResNet  and examine their performances of classifying images of EMINIST[6] on these different CNN infrastructures using multi-class cross-entropy loss and micro f1 score.

\section{Related Work}

\subsection{SGD + momentum}

SGD with momentum is a way to address the problem SGD has with local optima. By adding a momentum term determined by the previous gradient, SGD will be accelerated in the relevant direction, so it is easier to overcome local optima to search for the global, and converge faster than a normal SGD.[3]
\subsection{Conventional Adam}

Adam is a replacement optimization algorithm for stochastic gradient descent for training deep learning models. Adam combines the best properties of the Momentum and RMSProp[10] algorithms to provide an optimization algorithm with an adaptive learning-rate that can handle sparse gradients on noisy problems.[4]

\subsection{AdaBelief}
AdaBelief is a modified version of Adam, it utilizes "belief" to automatically select the proper step size at each gradient update. The algorithm claimed to have two properties, fast convergence as in adaptive gradient methods, and good generalization as in the SGD family.[1]

\subsection{Padam}
Padam (partially adaptive momentum estimation) [5] is a modified version of Adam. It tries to close the generalization gap of adaptive gradient methods by introducing a partially adaptive parameter which also resolves the “small learning rate dilemma”(initial learning rate for adaptive methods often small) for adaptive methods and allows for faster convergence[5]. Padam is showed empirically that it achieves the fastest convergence speed while generalizing as well as SGD with momentum[5].

\section{Methods and Algorithm}

We evaluate the performances of these different optimization algorithms empirically with  EMNIST Dataset[6].  AlexNet[8] and simplified versions of VGG[7], ResNet[9] are trained by Algorithm \ref{TrainModel} on the training set using all optimization methods described above with the same initialization. We set learning rate for all optimization algorithms to 0.0005 and the rest of the hyper-parameters default except for SGD we have momentum=0.9. All of the models are trained for 30 epochs with batch size of 64. Given imbalanced labels, performances are evaluated empirically using micro f1 score. For all models, we modify the input channel to be 1 (all samples have only one channel) and output dimension to be 47. All experiments are conducted using Pytorch[2]. Due to the time constraint and computational cost, all experiments are conducted for only one random seed. 


\begin{algorithm}  
\caption{Training A Model With An Optimizer}
\label{TrainModel}
\begin{algorithmic}[1] 
\REQUIRE $d$ is the data of EMNIST; $epochSize$ is the epoch size; $batchSize$ is the batchSize; $opt$ is Optimizer Algorithm; $model$ is the training Model; 
\STATE $(trainSet, validationSet) \gets $ split($d$)  
\FOR {each epoch $e=1,2,...,epochSize$}  
\STATE $(Source_1, Target_1), (Source_{2}, Target_{2}), ... \gets $(split $trainSet$ in equal parts of batchSize) 
\FOR {each batch $b=1,2,...,batchSize$}  

\STATE $prediction \gets TrainModel(model, Source_b, Target_b)$
\STATE $loss \gets CalculateCrossEntropyLoss(prediction, Target_b)$
\STATE $PerformBackwardPropagation(loss)$
\STATE $UpdateModelParameters(opt)$
\STATE $ClearUpGradient(opt)$
\ENDFOR
\ENDFOR

\end{algorithmic}  
\end{algorithm}  

\begin{figure}[htp]
    \centering
    \includegraphics[width=13.5cm]{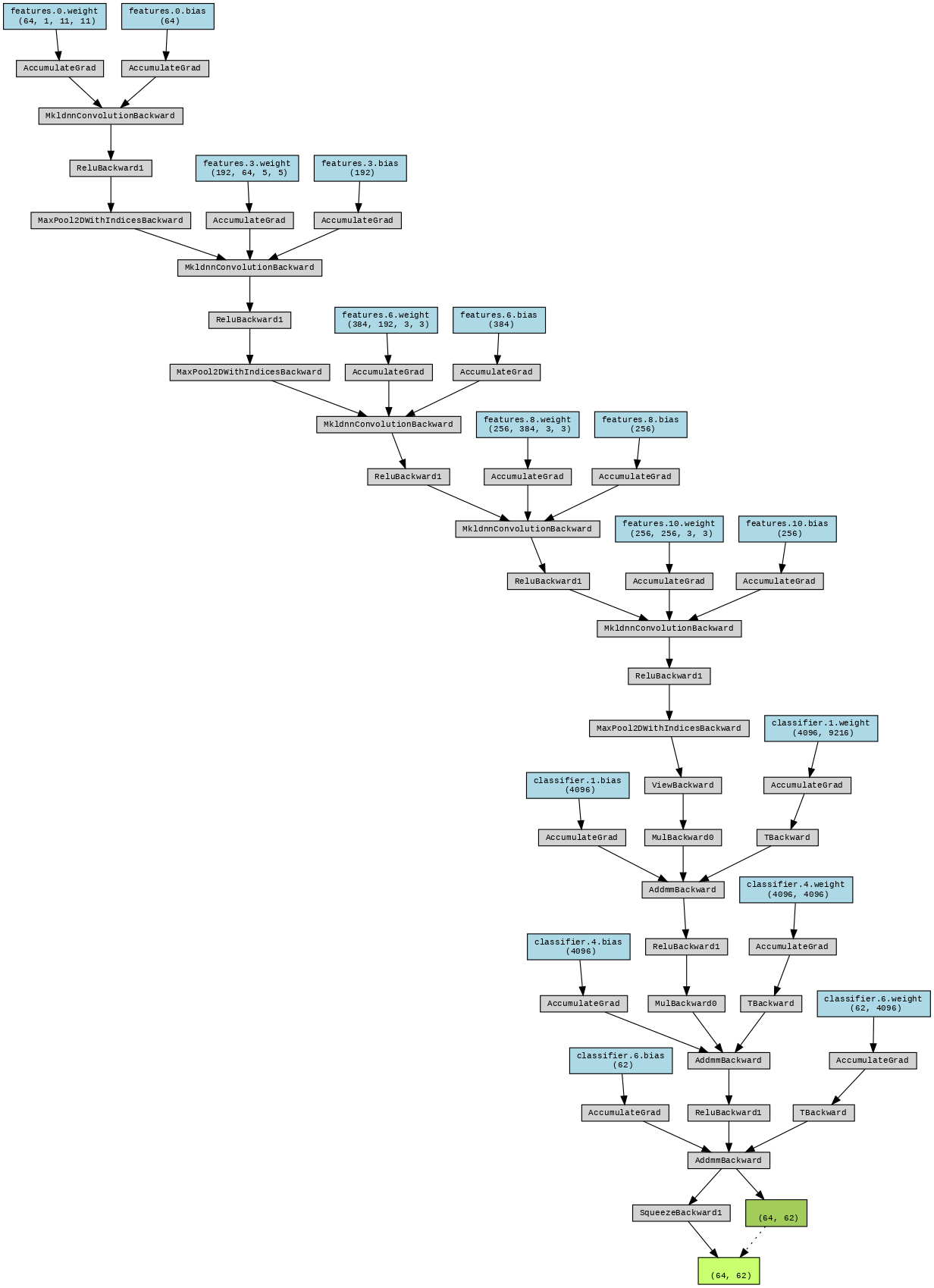}
    \caption{AlexNet Structure}
    \label{alexnet}
\end{figure}

\begin{figure}[htp]
    \centering
    \includegraphics[width=15cm, height=20cm]{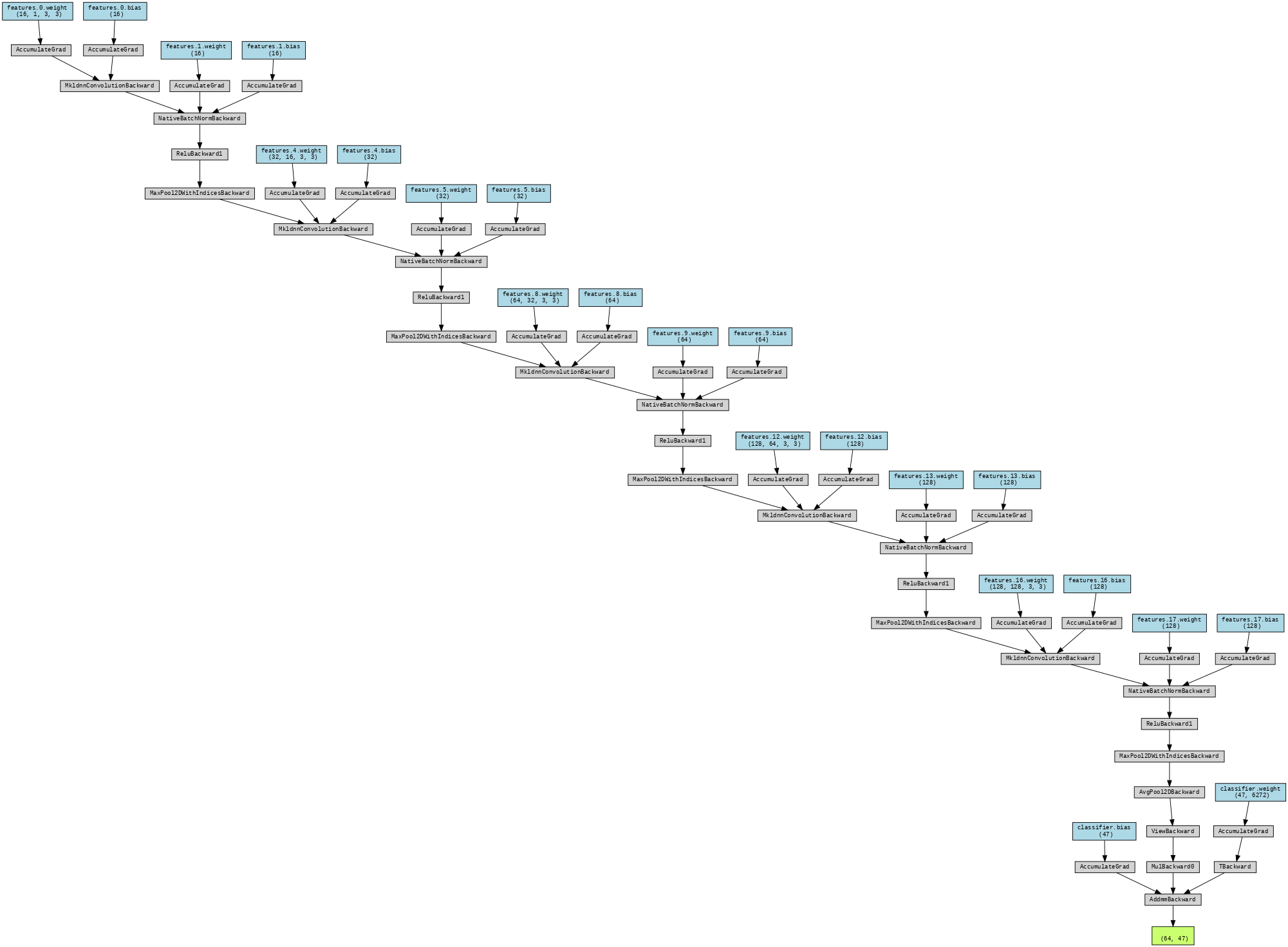}
    \caption{VGG Structure}
    \label{vgg_structure}
\end{figure}

\begin{figure}[htp]
    \centering
    \includegraphics[width=15cm]{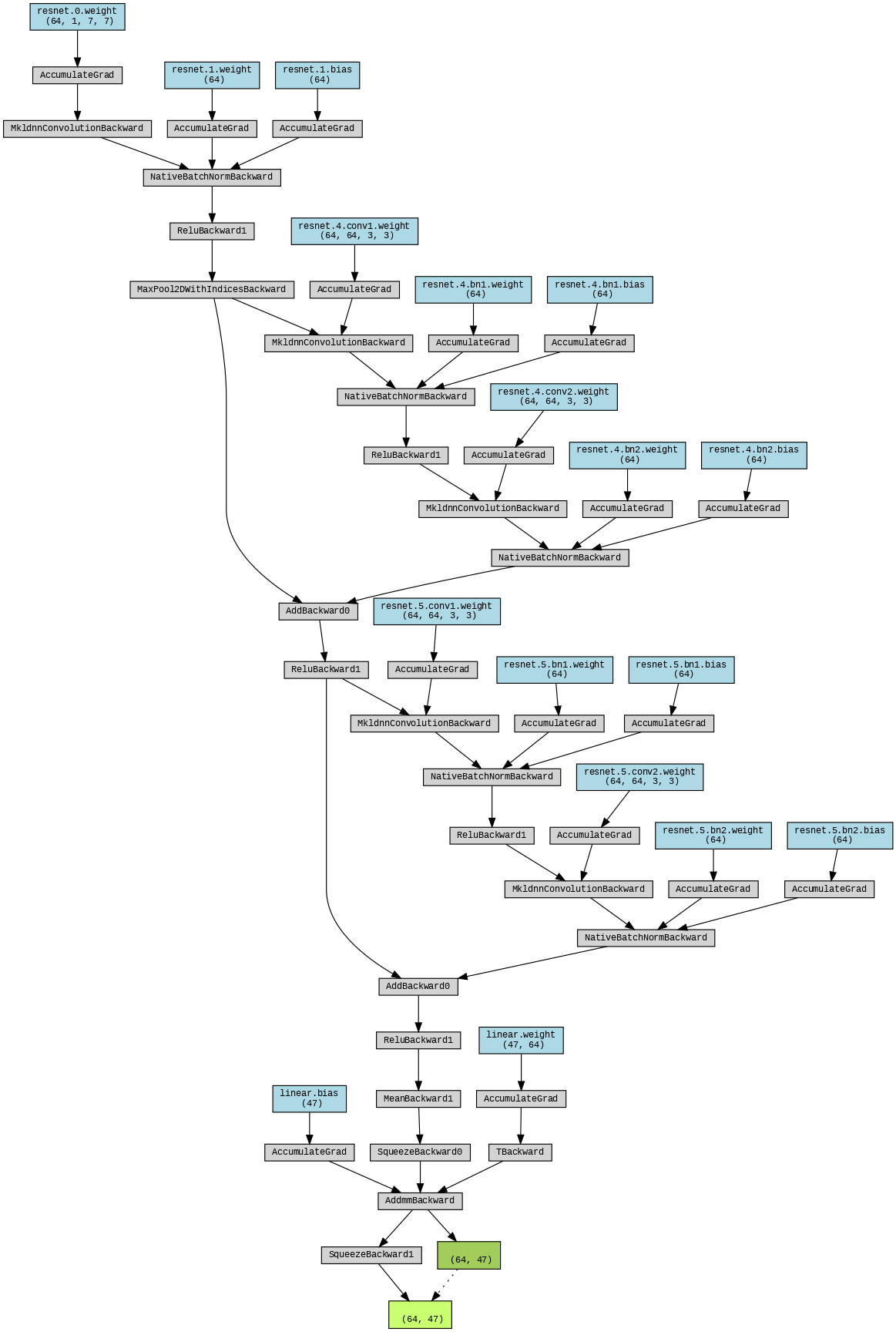}
    \caption{ResNet Structure}
    \label{resnet}
\end{figure}

We apply the following 3 kinds of models to modify and train from scratch :

\begin{itemize}
    \item \textbf{AlexNet}:  We use full AlexNet[8].(as shown in Figure \ref{alexnet})
    \item \textbf{VGG}: We shrink the standard VGG11[7] model to avoid overfit. In each of 5 blocks, we only leave one convolutional layer and one max pool layer. We also reduce the kernel number in each layer to $\frac{1}{4}$ (as shown in Figure \ref{vgg_structure}).
    \item \textbf{ResNet}: We only use the first layer of the ResNet18[9] (as shown in Figure \ref{resnet}), as the original ResNet18 structure is too heavy for this task (overfit the training set at epoch 2) and too computationally expensive. 
\end{itemize}




\section{Experiments and Results}

\begin{table}[h]
    \centering
    \begin{tabular}{lllllll}
        \toprule
        Optimizers & \multicolumn{2}{c}{AlexNet} &  \multicolumn{2}{c}{VGG} &  \multicolumn{2}{c}{ResNet}\\
        \cmidrule{2-3} \cmidrule{4-5} \cmidrule{6-7}\\
        {} & Best Test f1 Score & Epoch \# & Best Test f1 Score   & Epoch \# & Best Test f1 Score  & Epoch \#\\
        \midrule
        SGD & 0.90106  & 30 & 0.90202  & 26 & 0.90450 & 22\\
        Adam  &  0.90397 & 11 & 0.90267 & 19 & 0.89720 & 5\\
        AdaBelief  & 0.90388 & 12 & 0.90278 & 23 & 0.89873 & 16 \\
        Padam  & 0.90591 & 14 &  0.90196 & 30 & 0.90282 & 30 \\
        \bottomrule
    \end{tabular}
    \caption{Best Validation f1 scores and corresponding epoch numbers of them for each combination of models and optimizers on EMNIST dataset}
    \label{models_results}
\end{table}

\subsection{AlexNet}
\begin{figure}[htp]
    \flushleft
    \includegraphics[scale=0.3]{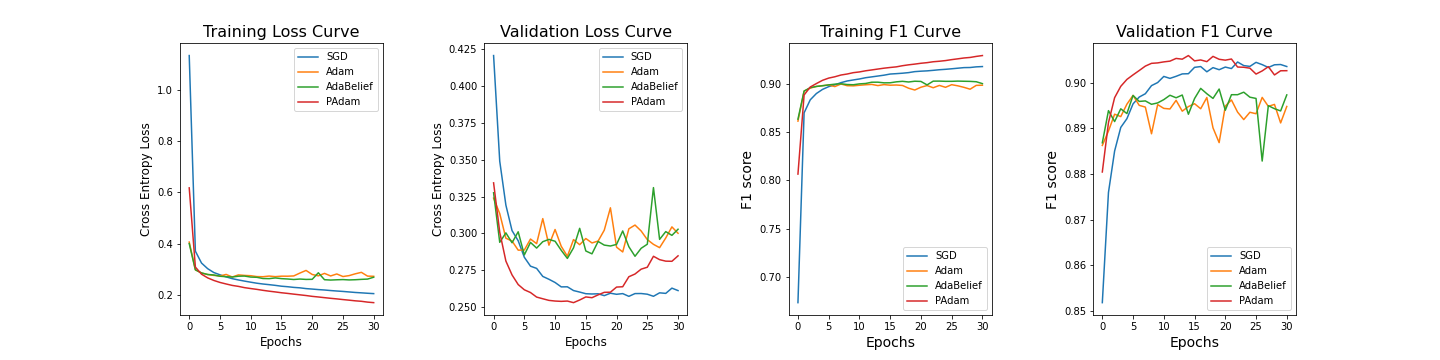}
    \caption{AlexNet Evaluation}
    \label{alexnet_images}
\end{figure}

From Table \ref{models_results}, we find out that Padam[5] reached the best f1 score in test data at epoch 14. As shown in Figure \ref{alexnet_images}, we can observe that Adam[4] and AdaBelief[1] optimizer begin to show the sign of overfitting at epoch 3, their train losses are flattened and test losses start to fluctuate. On the other hand, train losses for both SGD and Padam smoothly decrease for all 30 epochs, however, start from epoch 14, Padam begins to overfit quickly, the test loss begins to increase, while test loss for SGD just start to be flattened, as shown in f1 score curves (Figure \ref{alexnet_images}).Therefore, in this task, Padam achieved the best performance at epoch 14. However, it shows sign of overfitting in the test loss and test f1 score curve, whereas SGD with Momentum keep slowly increasing performance. SGD with Momentum may perform best after it converges.

\subsection{VGG}

\begin{figure}[htp]
    \flushleft
    \includegraphics[scale=0.3]{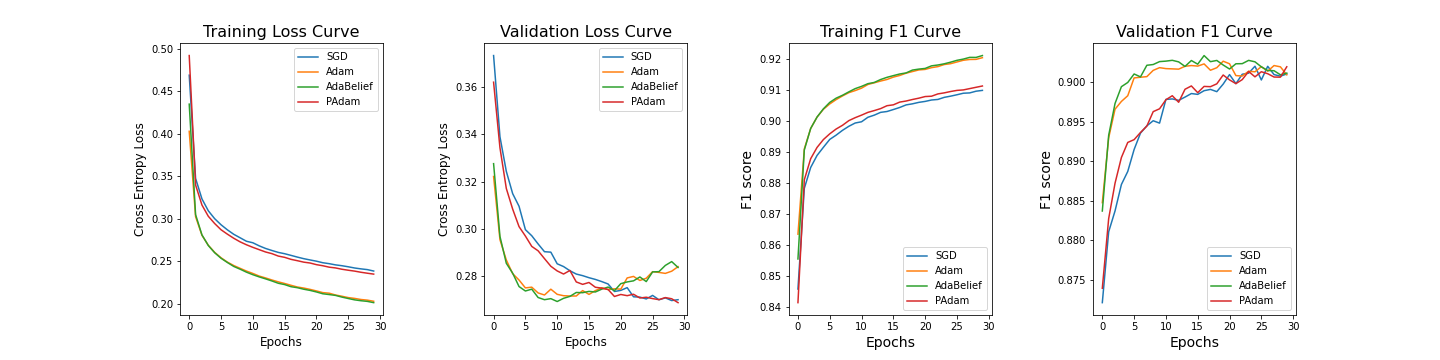}
    \caption{VGG Evaluation}
    \label{vgg_images}
\end{figure}

 As shown in Figure \ref{vgg_images}, the test performance of model is not stable. It is observed that the trends for Adam and AdaBelief,  Padam and SGD with Momentum show pairwise similarity. In the beginning, Adam and AdaBelief perform better than SGD with Momentum and Padam. And then they are surpassed due to overfit.In epoch 30, test loss of SGD with Momentum and Padam is less than the smallest loss of Adam and AdaBelief before Adam and AdaBelief begin to show the sign of overfit. On the other hand, although among 30 epochs the best test f1 score is achieved by AdaBelief (as shown in Table \ref{models_results}), SGD with Momentum or Padam may surpass it in the following epochs. Therefore, Padam or SGD with Momentum may perform best after it converges

\subsection{ResNet}

\begin{figure}[htp]
    \flushleft
    \includegraphics[scale=0.3]{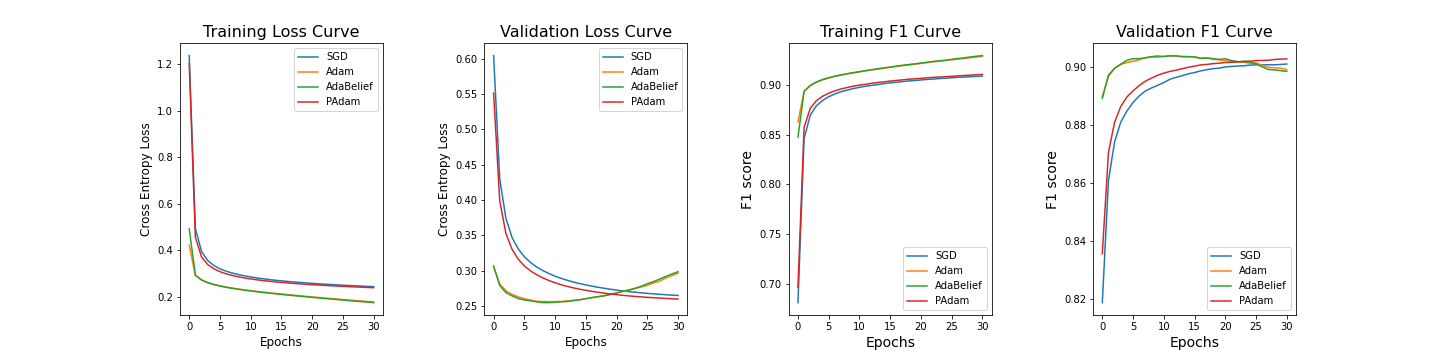}
    \caption{ResNet Evaluation}
    \label{res_images}
\end{figure}

From table \ref{models_results} we can see that the best test f1 score in 30 epochs is achieved using Adam. In the training and test loss plot (figure \ref{res_images}), we observe that in general the trends for Adam and AdaBelief,  Padam and SGD with Momentum show pairwise similarity.And Padam shows a slightly advantage of test loss over SGD with Momentum. Even though Adam has the best test f1 score at epoch 11, both Adam and AdaBelief start to show signs of overfitting on the test loss plot. On the other hand, the test losses for both Padam and SGD + Momentum decrease smoothly. The results above are consistent across f1 scores (figure \ref{res_images}). Hence, we can conclude that in this task, the local optima is achieved using Adam, but Adam and AdaBelief show signs of overfitting whereas Padam and SGD with Momentum leans smoothly, it is possible that Padam and SGD with Momentum outperform Adam and AdaBelief after they converge.

\section{Discussion  and  Conclusion}

From the experiments, we can conclude that Adam and AdaBelief have similar performance on this task, at the same time, Padam and SGD + Momentum have similar performance on this task. In general, the learning curves for SGD and Padam are much smoother than the learning curves for Adam and AdaBelief. In most cases, Adam and AdaBelief reach the local optima within 30 epochs but start to overfit right after, whereas Padam and SGD + Momentum leans smoothly, it is possible that Padam and SGD + Momentum outperform Adam and AdaBelief after they converge. Besides the results, the limitations in this project are obvious. Firstly, the results are not averaged over multiple random seeds because of the computational cost and time constraint. This problem can be solved by run multiple random trails with different random seeds and average the results over them. Secondly, the models we use in this project are modified versions of the original models, the experiment results may not be reproducible when we have the complete models. Lastly, we only run this experiment with limited epochs and one set of hyperparameters, our results may be local and may not carry across a different set of hyperparameters.

\section{Contributions}


\begin{itemize}
    \item \textbf{Zhaoyang Zhu}: benchmark algorithm propetyping, training ResNet from scratch
    \item \textbf{Haozhe Sun}: benchmark algorithm propetyping, training AlexNet from scratch
    \item \textbf{Chi Zhang}:  benchmark algorithm propetyping, training VGG from scratch
\end{itemize}

\section{References}

\small
[1] Juntang Zhuang, Tommy Tang, Yifan Ding, Sekhar Tatikonda,  Nicha Dvornek1,
Xenophon Papademetris\ \& James S. Duncan1\ (2020) AdaBelief Optimizer: Adapting Stepsizes by the Belief in Observed Gradients. 34th Conference on Neural Information Processing Systems (NeurIPS 2020), Vancouver, Canada.

[2] Paszke, A. et al. Pytorch: an imperative style, high-performance deep learning library. In Advances in Neural Information Processing Systems 32 (eds Wallach, H. et al.) 8024–8035 (Neural Information Processing Systems, 2019).

[3] Sebastian Ruder. An overview of gradient descent optimization algorithms. https://ruder.io/optimizing-gradient-descent/index.html

[4] Diederik P. Kingma\ \& Jimmy Ba\ (2015) ADAM: A METHOD FOR STOCHASTIC OPTIMIZATION. arXiv preprint arXiv:1412.6980v9

[5] Jinghui Chen and Quanquan Gu. Closing the generalization gap of adaptive gradient methods in
training deep neural networks. arXiv preprint arXiv:1806.06763, 2018.

[6] Gregory Cohen, Saeed Afshar, Jonathan Tapson, and Andre van Schaik. EMNIST: an extension of MNIST to ´
handwritten letters. arXiv preprint arXiv:1702.05373, 2017.

[7] K. Simonyan and A. Zisserman, “Very deep convolutional networks for large-scale image recognition,” in International Conference on Learning Representations (ICLR), 2015.

[8] A. Krizhevsky, I. Sutskever, and G. E. Hinton.
ImageNet classification with deep convolutional
neural networks. In Proceedings of NIPS, pages
1106–1114, 2012. papers.nips.cc/paper/4824-
imagenet-classification-with-deep-convolutionalneural-networks.pdf.

[9] K. He, X. Zhang, S. Ren, and J. Sun. Deep
residual learning for image recognition. In
Proceedings of CVPR, pages 770–778, 2016.
arxiv.org/abs/1512.03385.

[10] Tieleman T and Hinton E 2012 Lecture 6.5 - rmsprop, COURSERA: Neural networks for
machine learning
\newpage
\section*{Appendix}

\end{document}